**Title:**
Scalable predictive processing framework for multitask caregiving robots

**Authors:**
Hayato Idei,[1*] Tamon Miyake,[2] Tetsuya Ogata,[3] Yuichi Yamashita[1*]

**Affiliations:**
[1]Department of Information Medicine, National Institute of Neuroscience, National Center of Neurology and Psychiatry；Tokyo, Japan.
[2]Future Robotics Organization, Waseda University; Tokyo, Japan
[3]Department of Intermedia Art and Science, Waseda University; Tokyo, Japan.
*Corresponding author. Email: idei@ncnp.go.jp or yamay@ncnp.go.jp

**Abstract:**
The rapid aging of societies is intensifying demand for autonomous care robots; however, most existing systems are task-specific and rely on handcrafted preprocessing, limiting their ability to generalize across diverse scenarios. A prevailing theory in cognitive neuroscience proposes that the human brain operates through hierarchical predictive processing, which underlies flexible cognition and behavior by integrating multimodal sensory signals. Inspired by this principle, we introduce a hierarchical multimodal recurrent neural network grounded in predictive processing under the free-energy principle, capable of directly integrating over 30,000-dimensional visuo-proprioceptive inputs without dimensionality reduction. The model was able to learn two representative caregiving tasks, rigid-body repositioning and flexible-towel wiping, without task-specific feature engineering. We demonstrate three key properties: (i) self-organization of hierarchical latent dynamics that regulate task transitions, capture variability in uncertainty, and infer occluded states; (ii) robustness to degraded vision through visuo-proprioceptive integration; and (iii) asymmetric interference in multitask learning, where the more variable wiping task had little influence on repositioning, whereas learning the repositioning task led to a modest reduction in wiping performance, while the model maintained overall robustness. Although the evaluation was limited to simulation, these results establish predictive processing as a universal and scalable computational principle, pointing toward robust, flexible, and autonomous caregiving robots while offering theoretical insight into the human brain's ability to achieve flexible adaptation in uncertain real-world environments.

**Main Text:**

INTRODUCTION

As societies worldwide age rapidly, the growing demand for long-term care is exacerbated by an increasingly severe shortage of professional caregivers (*1–3*). Physically demanding tasks such as patient repositioning or body cleaning are not only labor intensive but also a leading cause of musculoskeletal disorders, particularly lower-back pain, among caregivers (*4, 5*). To address these challenges, various assistive robotic technologies have been developed (*6–12*), ranging from transfer devices and exoskeletons to humanoid systems designed for lifting or repositioning patients. However, most existing systems are either intended to support human operators or are specialized for a single, narrowly defined task, thus limiting their utility across the diverse and unpredictable scenarios encountered in real care settings. Therefore, there is an urgent need for next-generation care robots capable of acting autonomously and flexibly across multiple caregiving tasks. However, this remains a formidable challenge as such tasks involve direct interaction with humans and are characterized by substantial uncertainty and variability. Moreover, caregiving behaviors, such as supporting body parts, adjusting posture, or wiping body surfaces, differ fundamentally in their motor patterns, interaction objects, and sources of uncertainty. The development of a unified computational framework that enables a single robot to learn and generalize across these heterogeneous caregiving tasks represents an open and pressing frontier in robotics.

By contrast, humans can flexibly perform a wide range of caregiving tasks with a single brain, even under uncertain and dynamic conditions. Such cognitive flexibility is thought to arise from the brain's remarkable capacity to efficiently process the continuous influx of massive, multimodal sensory signals and to contextualize them for adaptive behavior. An influential theoretical framework in neuroscience is predictive processing, a computational framework under the free-energy principle that explains cognition as the minimization of prediction errors between sensory inputs and internally generated predictions (*13–18*). Rather than passively receiving inputs, the brain is viewed as an active inference system that continuously anticipates sensory signals and updates its internal models in response to prediction errors. This perspective has increasingly inspired research in artificial intelligence and robotics, where predictive processing–based neural architectures are being explored as a means of endowing robots with adaptive, brain-like learning capabilities across diverse tasks (*19–28*).

Despite this promise, a fundamental challenge remains: extending predictive processing to the direct integration of high-dimensional, multimodal sensory data, which is a prerequisite for robust multitask learning in real-world robotics. Current robotic approaches often rely on preprocessing, dimensionality reduction, or additional external processing modules, such as specialized transformations or attention-based mechanisms, to extract features from each modality (e.g., vision or proprioception), typically in a task-specific manner. Although effective in constrained or well-defined settings, such strategies inherently limit generalization because they depend on modality- and task-specific assumptions. This limitation becomes particularly problematic in multitask caregiving scenarios, where tasks differ widely in motor patterns, sensory targets, and sources of uncertainty. However, a unified computational framework that can directly integrate raw, high-dimensional, and uncertain sensory signals across modalities is lacking. Such a framework would not only enable robust and flexible learning across heterogeneous caregiving tasks but also validate predictive processing as a general computational principle for intelligent behavior under real-world uncertainty.

Here, we introduce a scalable predictive-coding–inspired variational recurrent neural network (PV-RNN) designed to integrate high-dimensional, multimodal sensory information. PV-RNN is a generative model that infers probabilistic and dynamic latent states under the free-energy principle, thereby capturing hidden stochastic structures in temporal data and remaining robust even in continuous and uncertain time series arising from human–robot interactions (*29–31*). In our earlier work, we extended PV-RNN into a hierarchical multimodal architecture and showed that it could self-organize dynamic modulations of hierarchical uncertainty estimation, while accounting for complex neural phenomena such as sensory attenuation and allostasis, as reported in neuroscience (*32–33*). Building on this foundation, we refined the framework to enhance predictive performance/scalability by introducing additional feedforward processing, enabling the model to better handle high-dimensional sensory signals with rapid, jagged temporal variations, while still operating solely under the principle of prediction error minimization (Fig. 1A and 1B). In contrast to conventional methods that depend on handcrafted features or dimensionality reduction, the proposed scalable PV-RNN learns end-to-end through variational free-energy minimization, which directly integrate raw, high-dimensional sensory signals across modalities. This capability enables the model to flexibly self-organize latent representations suited to diverse modalities and tasks, thereby offering a unified framework for multitask learning in robotics. Beyond its engineering contributions, evaluating the scalable PV-RNN on caregiving tasks also provides scientific value, enabling us to test whether prediction error minimization, as a single computational principle, can support the dynamic processing of high-dimensional multimodal information and yield robust adaptive behavior under uncertainty, as observed in humans.

## RESULTS

### Task description

This study investigated robot learning in caregiving tasks through high-dimensional visuo-proprioceptive integration. We employed the Dry-AIREC humanoid robot developed by Tokyo Robotics Inc., Tokyo, Japan (Fig. 1C). Dry-AIREC is equipped with multiple sensing modalities, including proprioceptive and visual sensors, providing a suitable platform for studying embodied interaction. The head was fitted with binocular RGB cameras, and each arm had seven degrees of freedom (DOFs) with torque sensors at every joint. For input to the scalable PV-RNN, we used 28-dimensional proprioceptive signals (14 joint angles and torques) and 32,256-dimensional visual signals (binocular RGB images at 16×16, 32×32, and 64×64 resolutions).

The robot was positioned in front of a bed on which a mannequin (height: 1800 mm; shoulder width: 470 mm; leg length: 885 mm; arm length: 750 mm; weight: 8 kg) served as the care recipient. We focused on two representative caregiving tasks, repositioning and wiping, that are central to maintaining the health and quality of life of individuals with limited mobility (Fig. 1D). The repositioning task (supine-to-sitting transfer) comprised three subtasks: (i) reaching the left hand to the mannequin's neck (reaching), (ii) placing the left hand under the upper back while the right hand supported the legs (holding), and (iii) lifting the upper body (lifting). The wiping task also comprised three subtasks: (i) reaching for a towel placed on the mannequin (reaching), (ii) wiping the body surface (wiping), and (iii) completing the wiping motion (releasing).

The learning and test data were obtained through teleoperation using a motion-capture system under three bed heights (590, 650, and 710 mm). We recorded visuo-proprioceptive sequences for 18 learning trials (9 per task, 3 per height) and 6 test trials (3 per task, 1 per height).

The repositioning task required dynamic manipulation of a rigid but heavy body with shifting load distribution, while adapting to variations in bed height, initial posture, and motion timing. By contrast, the wiping task required manipulation of a flexible towel across curved and uneven surfaces, while also accommodating variations in bed height, towel position, initial posture, and motion pattern. Both tasks required accurate target reaching, adaptation to object properties, and robustness to diverse spatiotemporal uncertainties. Together, they represent qualitatively distinct sources of variability and rigid-body dynamics under load versus flexible-object surface interaction. The central question addressed in this study was whether a single computational model can simultaneously learn and generalize across heterogeneous caregiving tasks.

**Predictive processing of the neural network**

The scalable PV-RNN is grounded in the free-energy principle; rather than directly consuming sensory inputs, it learns and infers by minimizing variational free energy across time steps (*32*).

$$F_t = \underbrace{\frac{1}{2}(x_t - \hat{x}_t)^2}_{\text{Prediction error}} + W \sum_{l=1}^{3} \underbrace{D_{KL}\left[q(z_t^{(l)}|e_{t:T}) || p(z_t^{(l)}|d_{t-1}^{(l)})\right]}_{\text{Influence of prior beliefs}} \quad (1)$$

Here, $x_t$ denotes the observed sensory inputs, $\hat{x}_t$ the predicted sensory inputs, $z_t^{q,(l)}$ the posterior belief (latent states inferred from observations), $z_t^{p,(l)}$ the prior belief (latent states predicted from past experience), $e_{t:T}$ the back-propagated error signal, and $d_t^{(l)}$ the states of recurrent units (time step $t$, last step $T$, network level $l$). Variational free energy has two components, the first quantifies sensory prediction error, while the second is the Kullback–Leibler divergence (KLD) between posterior and prior, capturing how past experience constrains posterior inference. The balancing hyperparameter $W$ was set to 0.005, following our prior work on multimodal PV-RNNs (*32*).

During learning, both the time-invariant synaptic weights $\omega$ and the sequence-specific posterior beliefs $z_{0:T}^{q,(l)}$ are updated to minimize the cumulative free energy across entire learning sequences. Importantly, posterior beliefs are updated independently for each sequence, enabling the simultaneous learning of multiple tasks.

During testing, the synaptic weights remain fixed, and the model performs online inference by updating $z_{t-H+1:t}^{q,(l)}$ within a sliding short-time window ($H = 30$) at each time step. This procedure locally minimizes the variational free energy, thereby implementing an adaptive inference process that dynamically adjusts the latent states to incoming sensory information.

Thus, the proposed scalable PV-RNN functions entirely through predictive processing based on variational free-energy minimization, without requiring handcrafted feature extraction, dimensionality reduction, or any additional external processing. For evaluation, we trained five scalable PV-RNNs with different random initializations to reconstruct visuo-proprioceptive sequences from 18 learning datasets. The trained networks were then tested in simulation-based generalization experiments using six test datasets, with five inference trials per sequence, yielding 30 test trials per trained network. All test evaluations were conducted in a simulation rather than on a physical robot.

**Experiment 1: Multitask generalization through hierarchical probabilistic inference**

Figure 2A illustrates an example of successful online inference for a test visuo-proprioceptive sequence in the repositioning task. Figure 2B presents results of an ablation analysis quantifying how latent states in each network module contributed to visual predictions. Contribution strength is visualized in grayscale, with brighter (whiter) regions indicating image areas more strongly influenced by the ablated module (see Materials and Methods). Figure 2B enables comparison between the grayscale visualization of visual contributions from each module and the actual image, where contours are highlighted. These results revealed self-organized hierarchical task representations. The exteroceptive module maintained continuous visual representations, particularly highlighting regions corresponding to the robot's left arm and the mannequin's face. The multimodal associative module exhibited strong latent dynamics when the visuo-proprioceptive relationship was most prominent, for instance, as the robot's hands approached the mannequin (steps 0–75) or during the lifting phase (steps 200–300). In contrast, the executive module showed marked changes in latent states and enhanced visual representations specifically at subtask boundaries, for example, at the onset of reaching (step ~0), at the initiation of left- and right-hand movements during the holding phase (steps ~75 and ~125), and at the beginning of the lifting phase (step ~200). These patterns suggest that the executive module functioned as a controller for motion switching, selectively signaling transitions between distinct subtask segments.

Figure 3 shows the corresponding results of the wiping task. Overall, the three modules displayed similar hierarchical specializations as in the repositioning task. The exteroceptive module maintained continuous visual representations, particularly highlighting regions corresponding to the robot's left arm and the mannequin's body. The multimodal associative module captured dynamic visuo-proprioceptive integration, particularly during the reaching, wiping, and releasing phases (steps 75–225). The executive module again exhibited state changes at subtask transitions, strongly detecting the shifts from initial posture to reaching and from wiping to releasing, while the transition from reaching to wiping appeared smoother. Notably, in predicted images (Fig. 3B), the mannequin's body contours occluded by the robot arms were inferred, demonstrating the model's capacity to complete uncertain hidden states. Collectively, these findings indicate that scalable PV-RNN developed hierarchical processing, with each module assuming distinct roles in dynamically allocating task-dependent attention and controlling subtask transitions.

Finally, Figure 4 compares task-specific uncertainty dynamics estimated by the network. For the mean level of uncertainty (Fig. 4A), prior sigma values were first averaged across all time steps and latent states within each trial, and then further averaged across 75 test runs for each task (five trials for each of the three test sequences and five independently trained networks). For variability in uncertainty (Fig. 4B), we quantified temporal fluctuations of prior sigma by computing, at each time step, the absolute difference from its value at the previous step, averaging these differences across all time steps and latent states within each trial, and then averaging them across the 75 test runs. Using these aggregated measures, we found that the mean level of estimated uncertainty was similar between repositioning and wiping, whereas variability in uncertainty was significantly greater for the wiping task. This suggests that the model flexibly adapted to differences in volatility across tasks, enabling generalization despite qualitatively distinct sources of uncertainty.

**Experiment 2: Robustness under uncertainty through multimodal integration**

To evaluate whether our model maintains robustness under uncertain sensory conditions, we tested its performance when visual inputs were degraded in resolution. Figure 5A presents predicted 16×16 and 64×64 images when the model updated its latent states using only low-resolution (16×16) visual inputs in combination with proprioceptive inputs. For comparison across trials, results are aligned at the same time step, revealing variability in motion timing and patterns among sequences. Remarkably, the model was able to generate high-resolution (64×64) visual predictions even though these signals were not provided during inference, accurately capturing task-dependent differences in timing and motion variability.

Figure 5B quantifies the effect of reduced visual resolution on the prediction error for full-resolution images. Values were first averaged across all time steps and sensory dimensions within each trial, and then further averaged across 150 trials (five trials for each of six test sequences and five networks). In both the presence and absence of proprioceptive input, performance decreased with lower visual resolution; however, this decline was significantly mitigated when proprioceptive input was available (red vs. blue bars in Fig. 5B).

These findings highlight the importance of multimodal predictive processing in ensuring robustness under both partial and uncertain sensory conditions. Once trained, the model can compensate for missing or degraded visual inputs by leveraging proprioceptive information, thereby sustaining accurate predictions. Moreover, the ability to operate with reduced-dimensional sensory inputs after training indicates the potential of our framework to achieve computational efficiency without sacrificing robustness.

**Experiment 3: Interference between tasks in learning**

To investigate the influence of multitask learning on generalization, we examined the performance while varying the balance of learning data between the repositioning and wiping tasks. Figure 6A shows the prediction errors on test sequences of the repositioning task when the number of repositioning training sequences was fixed at nine, while the number of wiping training sequences was varied (0, 3, 6, and 9). Prediction error values were first averaged across all time steps and sensory dimensions, including both exteroceptive and proprioceptive domains within each trial, and then further averaged across 75 trials (five trials for each of the three test sequences and five networks). Figure 6A demonstrates that increasing the amount of wiping data did not significantly influence the generalization performance in repositioning. This suggests that learning the wiping task did not interfere with the acquisition of task-specific representations required for repositioning.

In contrast, Figure 6B shows that increasing the amount of repositioning training data led to a statistically significant rise in the mean prediction error for the wiping task; however, the effect was modest and not disruptive to overall generalization performance. These results indicate that the scalable PV-RNN maintained robust performance without a marked drop even when the number of learned tasks increased, while still suggesting an asymmetric pattern of task interference.

**DISCUSSION**

This study introduced a brain-inspired framework for integrating high-dimensional, multimodal sensory information through a unified computational principle of prediction error minimization, implemented in a multimodal PV-RNN. By operating directly on raw sensory signals without task-specific preprocessing or external modules, the framework enabled flexible generalization across qualitatively distinct caregiving tasks. Using repositioning and wiping as representative cases, we demonstrated three key properties, (i) self-organization of hierarchical latent dynamics that regulated task-dependent allocation of attention, captured variability in uncertainty, and enabled the inference of occluded states; (ii) robustness under uncertain sensory conditions through visuo-proprioceptive integration; and (iii) asymmetric interference in multitask learning, where the more variable wiping task exerted minimal influence on repositioning, whereas learning the repositioning task resulted in a statistically significant but modest degradation in wiping performance, yet the model maintained overall robustness. Taken together, these results highlight the potential of predictive processing as a foundation for multitasking robotic intelligence in caregiving.

Beyond the engineering significance, these findings provide a broader discussion on the computational principles of cognition (*34–37*). Predictive processing under the free-energy principle has been proposed as a unifying framework for perception and action. However, its capacity to process real-world, high-dimensional multimodal inputs has remained underexplored. Our results address this gap by demonstrating the computational validity of the principle in robotics, rather than strict biological fidelity. The model reproduced properties often observed in humans: hierarchical task representation (*38–43*), inference of occluded states, and reliance on proprioception to maintain robust object manipulation when visual information is ambiguous, such as in dimly lit environments. Moreover, the asymmetric interference between repositioning and wiping suggests that tasks with higher variability in uncertainty (such as wiping) may exert weaker influence on the learning of other tasks, whereas tasks with lower variability (such as repositioning) tend to have stronger effects. This asymmetry may be partly explained by differences in how variability in uncertainty shapes internal representations. Tasks with greater variability in uncertainty (such as wiping) may foster the formation of more flexible and generalized latent representations, enabling robust adaptation across contexts. In contrast, tasks with lower variability (such as repositioning) may lead to more specialized and rigid representations, which can interfere with the learning of other tasks. This perspective provides insight into human developmental learning processes, suggesting that exploratory behaviors such as early motor babbling are effective for acquiring diverse skills without interfering with the learning of other behaviors (*44–46*). While speculative, these parallels suggest that our approach provides a promising computational tool to link neural principles with adaptive behavior in complex environments.

From an applied robotics perspective, the proposed framework offers practical advantages in three respects, robustness, as it can generalize across heterogeneous tasks without handcrafted preprocessing; efficiency, as it can later operate with reduced sensory resolution to lower computational costs; and scalability, as the same principle can be extended to additional tasks and modalities. These features underscore predictive processing as a versatile and principled pathway toward autonomous care robots capable of addressing diverse real-world demands.

However, this study has several limitations. First, it was restricted to simulation-based evaluations using teleoperated data; real-time control experiments on physical robots were not conducted. Second, although effective, the online inference process is computationally demanding, running at approximately 3 Hz in our experimental computer environment when processing proprioceptive signals together with binocular 16×16 RGB inputs. Thus, it requires more efficient implementations for real-time operation. Third, the mannequin used as the care recipient was mechanically simplified (lighter and more stable than humans), reducing ecological validity. Addressing these limitations will require advances in real-time optimization, more realistic human–robot interaction scenarios, and scaling to broader caregiving contexts.

Despite these constraints, our findings demonstrate that a brain-inspired computational framework can endow robots with robustness and versatility reminiscent of human cognition. Progress toward real-time implementations and scaling to richer task sets holds promise for the development of next-generation care robots capable of supporting humans in diverse and uncertain environments. Such robots would not only alleviate caregiver burden but also shed light on the computational principles that enable biological intelligence to thrive in complex worlds.

## MATERIALS AND METHODS

### Experimental design

The aim of this study was to develop and evaluate a brain-inspired framework for robust multitask learning in caregiving contexts. Specifically, we tested whether a single computational principle, prediction error minimization, can support the integration of high-dimensional, multimodal sensory information and enable generalization across qualitatively different tasks. To this end, we employed a scalable PV-RNN, a hierarchical multimodal generative model designed to infer probabilistic latent states under the free-energy principle. Using this framework, we investigated whether a robot could simultaneously learn and generalize two distinct caregiving tasks, repositioning and wiping, that differ in their motor patterns, interaction objects, and sources of uncertainty. Importantly, the present study did not involve real-time physical execution, instead, all evaluations were conducted in simulation using visuo-proprioceptive data sequences recorded through teleoperation.

### Robot setup and data acquisition

Learning and test data were obtained by teleoperating a dual-arm humanoid robot with a motion-capture system. During task execution, proprioceptive signals (joint angles and torques from both arms) and exteroceptive signals (binocular RGB images) were recorded as time-series data. Each signal was normalized to the range –0.9 to 0.9 using its maximum and minimum values, consistent with the activation functions of the neural network. The two caregiving tasks were (i) repositioning, in which the robot reached toward and lifted the upper back of a mannequin, requiring the manipulation of a rigid body with varying load distribution; and (ii) wiping, in which the robot reached for and manipulated a towel to clean the mannequin's curved surface, requiring control of a flexible object under variable contact conditions. Both tasks were performed under variability in initial mannequin posture, bed height, and action timing, thereby introducing spatiotemporal uncertainty into the visuo-proprioceptive data.

In total, 24 visuo-proprioceptive sequences were collected, 18 sequences (9 per task, 3 per bed height) for learning and 6 sequences (3 per task, 1 per bed height) for testing. To evaluate robustness across different initializations, we trained five scalable PV-RNNs with independently randomized synaptic weights. For testing, each trained model performed online inference across all test sequences with five independent trials, yielding 30 test trials per model.

**Prediction generation of the scalable PV-RNN**

The scalable PV-RNN was organized into three hierarchical levels, consistent with our previous work. Level 1 encoded low-level exteroceptive and proprioceptive representations, Level 2 integrated these modalities into mid-level sensorimotor representations, and Level 3 controlled higher-level patterns of visuo-proprioceptive integration. To enhance the processing of high-dimensional sensory inputs, an additional feedforward neural network was inserted between Level 1 and the sensory prediction outputs, enabling more flexible handling of complex signals.

Prediction generation was performed in a top-down manner through the network hierarchy. The internal state $h_{t,i}^{(s)}$ and output $d_{t,i}^{(s)}$ of the $i$th deterministic recurrent units of the $s$th target sequence at time step $t$ is computed as

$$h_{t,i}^{(s)} = \frac{1}{\tau}\left(\sum_{j \in I_{Sd}} \omega_{ij} d_{t-1,j}^{(s)} + \sum_{j \in I_{Sz}} \omega_{ij} z_{t,j}^{(s)} + \sum_{j \in I_{Hd}} \omega_{ij} d_{t,j}^{(s)} + b_i\right) + \left(1 - \frac{1}{\tau}\right) h_{t-1,i}^{(s)} \quad (i \in I_{Ed}, I_{Pd}, I_{Ad}, I_{Cd})$$

(2)

$$d_{t,i}^{(s)} = \tanh\left(h_{t,i}^{(s)}\right) \quad (i \in I_{Ed}, I_{Pd}, I_{Ad}, I_{Cd}) \quad (3)$$

Here, $I_{Ed}$, $I_{Pd}$, $I_{Ad}$, and $I_{Cd}$ denote index sets of deterministic recurrent states in the exteroceptive, proprioceptive, multimodal associative, and executive modules, respectively. Similarly, $I_{Ez}$, $I_{Pz}$, $I_{Az}$, and $I_{Cz}$ denote index sets of probabilistic latent states. $I_{Sd}$, $I_{Sz}$, and $I_{Hd}$ denote recurrent states of the same module, latent states of the same module, and recurrent states of the higher-level module, respectively. $\omega_{ij}$ is the synaptic weight from the $j$th to the $i$th neuron; $z_{t,j}^{(s)}$ is the output of the $j$th latent (posterior) state at time step $t$; $\tau$ is the time constant of the recurrent unit; and $b_i$ is the bias of the $i$th recurrent unit. A smaller $\tau$ yields faster-changing dynamics, while a larger $\tau$ produces slower dynamics. The initial internal states of the recurrent units $h_{0,i}^{(s)}$ ($i \in I_{Ed}, I_{Pd}, I_{Ad}, I_{Cd}$) were set to 0 (thus, $d_{0,i}^{(s)} = 0$).

The latent variable **z** follows a multivariate Gaussian distribution with diagonal covariance, implying independence between $z_{t,i}^{(s)}$ and $z_{t,j}^{(s)}$ for $i, j \in I_{Ez}, I_{Pz}, I_{Az}, I_{Cz} \land i \neq j$. The prior distribution was parameterized from previous recurrent states in the same module:

$$p\left(z_{t,i}^{(s)}\right) = p\left(z_{t,i}^{(s)} | d_{t-1,j}^{(s)}\right) = \mathcal{N}\left(z_{t,i}^{(s)}; \mu_{t,i}^{(s),p}, \sigma_{t,i}^{(s),p}\right) \tag{4}$$

$$\mu_{t,i}^{(s),p} = \tanh\left(\sum_j \omega_{ij} d_{t-1,j}^{(s)}\right) \tag{5}$$

$$\sigma_{t,i}^{(s),p} = \exp\left(\sum_j \omega_{ij} d_{t-1,j}^{(s)}\right) \tag{6}$$

where $(i \in I_{Ez} \wedge j \in I_{Ed}) \vee (i \in I_{Pz} \wedge j \in I_{Pd}) \vee (i \in I_{Az} \wedge j \in I_{Ad}) \vee (i \in I_{Cz} \wedge j \in I_{Cd})$.

The posterior distribution was calculated as

$$q\left(z_{t,i}^{(s)} | e_{t:T^{(s)}}^{(s)}\right) = \mathcal{N}\left(z_{t,i}^{(s)}; \mu_{t,i}^{(s),q}, \sigma_{t,i}^{(s),q}\right) \quad (i \in I_{Ez}, I_{Pz}, I_{Az}, I_{Cz}) \tag{7}$$

$$\mu_{t,i}^{(s),q} = \tanh\left(a_{t,i}^{(s),\mu}\right) \tag{8}$$

$$\sigma_{t,i}^{(s),q} = \exp\left(a_{t,i}^{(s),\sigma}\right) \tag{9}$$

$$z_{t,i}^{(s)} = \mu_{t,i}^{(s),q} + \sigma_{t,i}^{(s),q} \times \epsilon, \quad \epsilon \sim \mathcal{N}(0,1) \tag{10}$$

Here, $a_t^{(s)}$ is the adaptive internal state of units parameterizing the posterior distributions. These adaptive variables $a_t^{(s)}$ were initialized from the corresponding prior distributions states before the training or inference process.

Two fully connected feedforward neural networks (FNN2L) for exteroceptive signals and one FNN1L for proprioceptive signals mapped recurrent states to predictions:

$$f_{t,I_{Ef}}^{(s)} = FNN2L\left(d_{t,I_{Ed}}^{(s)}\right) \tag{11}$$

$$f_{t,I_{Pf}}^{(s)} = FNN1L\left(d_{t,I_{Pd}}^{(s)}\right) \tag{12}$$

$$\hat{x}_{t,i}^{(s)} = \begin{cases} \tanh\left(\sum_{j \in I_{Ef}} \omega_{ij} f_{t,j}^{(s)}\right) & (i \in I_{Eo}) \\ \tanh\left(\sum_{j \in I_{Pf}} \omega_{ij} f_{t,j}^{(s)}\right) & (i \in I_{Po}) \end{cases} \tag{13}$$

Here, $I_{Ef}$ and $I_{Pf}$ denote index sets of units in the lower layer of the FNN2L and FNN1L, respectively. $I_{Eo}$ and $I_{Po}$ denote the output units for exteroceptive and proprioceptive predictions, respectively.

**Cost function of the scalable PV-RNN**

Variational free-energy $F_t^{(s)}$ in the PV-RNN is formulated as:

$$F_t^{(s)} = -\underbrace{\log p\left(x_t^{(s)}\big|d_t^{(s)}\right)}_{\text{Accuracy}} + W \sum_{l=1}^{3} \underbrace{\left(D_{KL}\left[q(z_t^{(l),(s)}|e_{t:T}^{(s)})||p(z_t^{(l),(s)}|d_{t-1}^{(l),(s)})\right]\right)}_{\text{Complexity}} \qquad (14)$$

The first term (negative accuracy term) corresponds to the negative log-likelihood. Assuming that each sensory state follows a Gaussian distribution with unit variance, this reduces to the squared prediction error between the observed sensations $x_t^{(s)}$ and predicted $\hat{x}_t^{(s)}$ sensations (constant term omitted):

$$-\text{Accuracy} = \sum_{i \in I_{\text{Eo}} \vee I_{\text{Po}}} \frac{1}{2}\left(x_{t,i}^{(s)} - \hat{x}_{t,i}^{(s)}\right)^2 \qquad (15)$$

In practice, the accuracy term was normalized by the dimensionality of each sensory modality.

The second term (complexity) is the KLD between posterior and prior latent distributions, quantifying the influence of prior beliefs. Assuming both distributions are multivariate Gaussians with diagonal covariance, the KLD can be expressed analytically as

$$\text{Complexity} = \sum_{i} \left( \log \frac{\sigma_{t,i}^{(s),p}}{\sigma_{t,i}^{(s),q}} + \frac{\left(\mu_{t,i}^{(s),p} - \mu_{t,i}^{(s),q}\right)^2 + \left(\sigma_{t,i}^{(s),q}\right)^2}{2\left(\sigma_{t,i}^{(s),p}\right)^2} - \frac{1}{2} \right) \qquad (16)$$

Here, $i \in I_{\text{Ez}}, I_{\text{Pz}}, I_{\text{Az}}, I_{\text{Cz}}$. This complexity term was also normalized by the latent dimensionality of each module.

During learning, both synaptic weights $\omega$ and adaptive variables $a_{0:T^{(s)}}^{(s)}$ were updated to minimize cumulative free energy:

$$F_{learning} = \sum_{s \in I_s} \sum_{t=1}^{T^{(s)}} F_t^{(s)} \qquad (17)$$

During online inference, synaptic weights were fixed, and only adaptive variables were updated. Free energy was minimized within a sliding short-time window $H$:

$$F_{inference} = \sum_{t'=t-H+1}^{t} F_{t'} \qquad (18)$$

Adaptive variables within the time window were iteratively updated based on this local free energy, with the window advancing at each time step.

Crucially, this formulation enables the scalable PV-RNN to operate solely through variational free-energy minimization, without relying on handcrafted feature extraction, dimensionality reduction, or auxiliary processing modules external to the free-energy principle.

**Hyper-parameter setting of the scalable PV-RNN**

For simplicity, the dimensions of recurrent variables in the exteroceptive, proprioceptive,

multimodal associative, and executive modules were set to $N_{Ed} = N_{Pd} = N_{Ad} = N_{Cd} = 30$. Latent variable dimensions were set to $N_{Ez} = 40$ and $N_{Pz} = N_{Az} = N_{Cz} = 30$. These values were determined in a preliminary study that searched for the minimal configuration enabling successful reconstruction across all trained networks. Specifically, we first set identical dimensions for recurrent and latent variables (10, 20, or 30) and then increased the number of latent variables in the exteroceptive module to 40, which was necessary for adequate reconstruction. The fully connected layer in the proprioceptive pathway comprised 40 units, while the exteroceptive pathway included two fully connected layers with 40 and 50 units, respectively.

To incorporate a temporal hierarchy, we implemented multiple timescales across modules. In the lower perceptual modules, time constants were set as $\tau_{Pd} = \tau_{Ed} = 2$ or $4$ (half of the recurrent units each). In the multimodal associative module, $\tau_{Ad} = 4$ or $8$, and in the executive module $\tau_{Cd} = 8$ or $16$. Thus, higher-level modules exhibited slower neural dynamics than lower-level ones, realizing a hierarchical temporal structure.

Synaptic weights were initialized with the Xavier method (*47*). Biases of deterministic recurrent variables were initialized and fixed to random values drawn from a Gaussian distribution $\mathcal{N}(0,10)$, following a previous study (*48*). During learning, synaptic weights and adaptive (posterior) variables were updated for $itr_{learning} = 100{,}000$ iterations using the rectified Adam optimizer (*49*) with learning rate $lr_{learning} = 0.001$, $\beta_1 = 0.9$, $\beta_2 = 0.999$.

In online inference, only adaptive variables were updated. At each sensorimotor time step $t$, they were updated $itr_{inference} = 50$ times for a short time window of length $H = 30$ (or $H = t \text{ if } t < 30$), with learning rate $lr_{inference} = 1.0$. Parameter settings were selected by systematically searching combinations of $lr_{inference} = \{1.0\} \times itr_{inference} = \{20, 30, 40, 50\} \times H = \{10, 20, 30, 40, 50\}$, choosing the minimal configuration that achieved successful reconstruction of the test data.

**Ablation study**

To analyze how latent states at different hierarchical levels contributed to visual predictions, we conducted an ablation study during the online inference tests. For each test sequence, predicted images were generated under two conditions: (i) the standard condition, with all latent states intact, and (ii) the ablation condition, with latent states of a specific module removed. Predictions were averaged across 25 trials (five test trials for each of the five independently trained networks). At each time step, the pixel-wise difference between the two conditions was computed. The averaged differences were visualized in grayscale, where brighter regions indicated visual areas most influenced by the ablated module. Figures 2B and 3B present the representative results from the right camera at 64×64 resolution, revealing distinct module-specific contributions to hierarchical task processing.

**Statistical analysis**

Paired t-tests were used to compare the mean and temporal changes in prior sigma (Fig. 4). To analyze the effects of proprioceptive input on visual prediction errors under different image-resolution conditions, a two-way repeated-measures aligned rank transform analysis of variance (ART-ANOVA) with Holm-corrected post hoc tests was performed (Fig. 5). For examining prediction errors under different data-balance conditions, one-way ART-ANOVAs followed by Holm-corrected multiple comparisons were conducted (Fig. 6). All statistical tests were two-tailed, and the significance level was set at $p < 0.05$. Because this study involved an unprecedented computational simulation, it was difficult to estimate effect sizes in advance; therefore, no statistical methods were used to pre-determine the sample size. Given the high reproducibility of the simulation, a minimum sample size of five networks was adopted. Even with this small sample, the analyses revealed clear statistical differences, suggesting that larger samples would not have substantially altered the main findings. It should be noted that, due to the limited sample size, these statistical tests were performed primarily as supportive quantitative analyses to complement the descriptive and comparative findings, rather than to draw definitive inferential conclusions. All data analyses were conducted using R software (version 4.5.1).

**Acknowledgments:**

    **Funding:**

        Japan Science and Technology Agency ACT-X No. JPMJAX24C2 (HI)

        Japan Science and Technology Agency Moonshot R&D No. JPMJMS2031 (YY, TO)

        Japan Science and Technology Agency Core Research for Evolutional Science and Technology No. JPMJCR21P4 (YY)

        AMED Multidisciplinary Frontier Brain and Neuroscience Discoveries (Brain/MINDS 2.0) No. JP24wm0625407 (YY)

    **Author contributions:**

        Conceptualization: HI, YY
        Methodology: HI
        Investigation: HI, TM
        Visualization: HI, YY
        Funding acquisition: HI, TO, YY
        Project administration: HI, YY
        Supervision: TO, YY
        Writing – original draft: HI
        Writing – review & editing: HI, TM, TO, YY

**Competing interests:** Authors declare that they have no competing interests.

**Figures:**

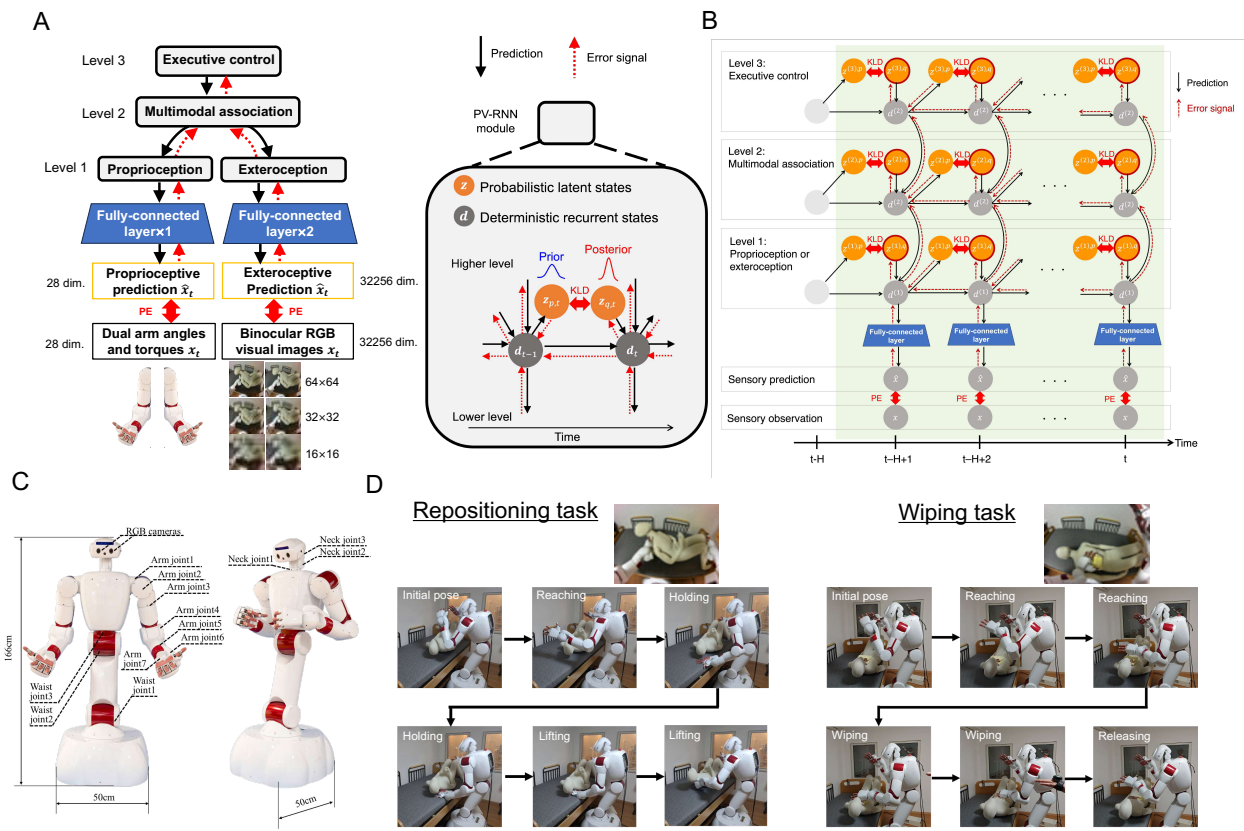

**Fig. 1. Computational framework and task setting.** (**A**) Architecture of the scalable PV-RNN. During learning, sequential posteriors of latent states across all modules and time-invariant synaptic weights are updated by minimizing variational free-energy accumulated over all training sequences. PE: prediction error; KLD: Kullback–Leibler divergence. (**B**) Temporal processing and online inference in the PV-RNN. During inference, posteriors in all modules are updated through minimization of variational free energy within a short time window ($H$), while synaptic weights remain fixed. For clarity, the distributed structure of the proprioceptive and exteroceptive modules is omitted. (**C**) The AIREC humanoid robot. (**D**) Two caregiving tasks used in this study: repositioning and wiping, both learned by the scalable PV-RNN.

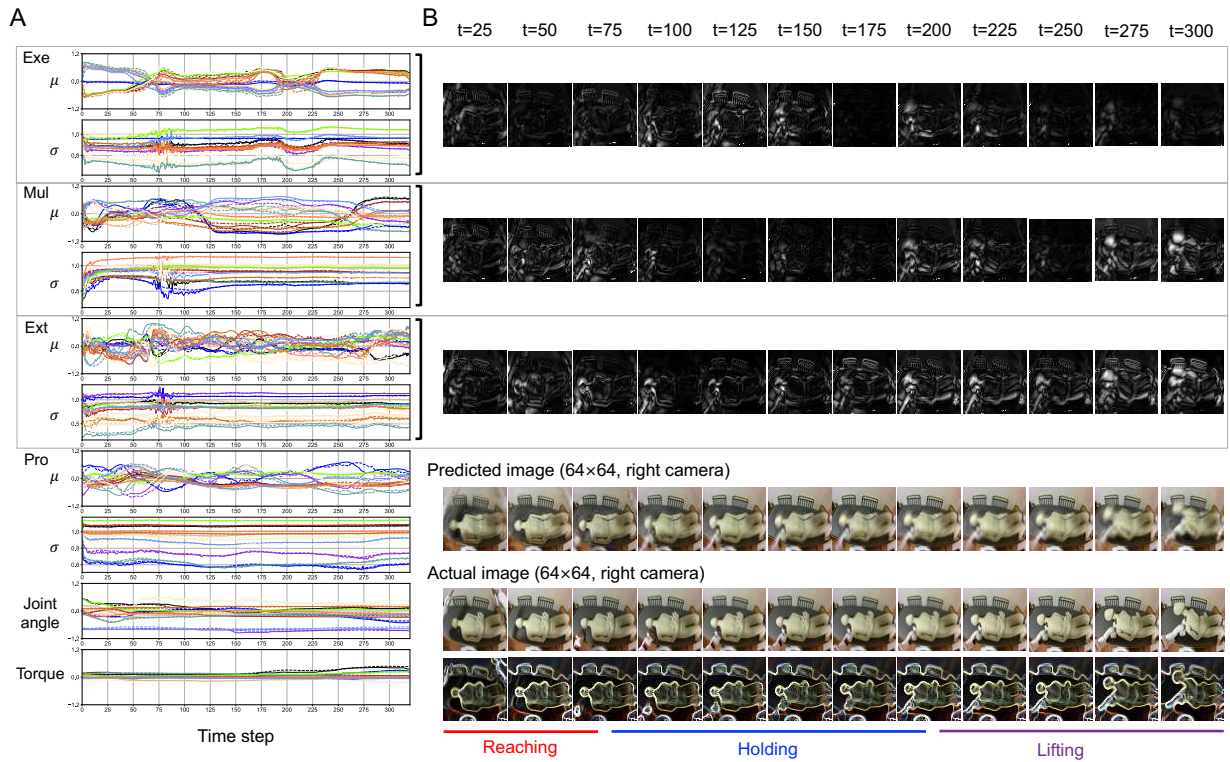

**Fig. 2. Hierarchical probabilistic inference during the repositioning task.** (**A**) Dynamics of probabilistic latent states in each module, along with predicted and actual proprioceptive inputs. For the mean $\mu$ and sigma $\sigma$ (standard deviation) of latent states, posteriors and priors are shown as solid and dashed lines, respectively. For joint angles and torques, actual and predicted values are indicated by solid and dashed lines, respectively. Different colors represent distinct latent states, joint angles, or torques. For clarity of presentation, the time series of latent states are plotted for 15 selected states. Exe: Executive module. Mul: Multimodal associative module. Ext: Exteroceptive module. Pro: Proprioceptive module. (**B**) Visual representations in each module, together with predicted and actual visual inputs. For reference, an additional image emphasizing the contours of the actual visual inputs is shown beneath them. Contributions of latent states were analyzed through an ablation study. Brighter regions indicate greater influence of the corresponding module's latent states on generating visual predictions. Results are shown for 64×64 images from the right camera.

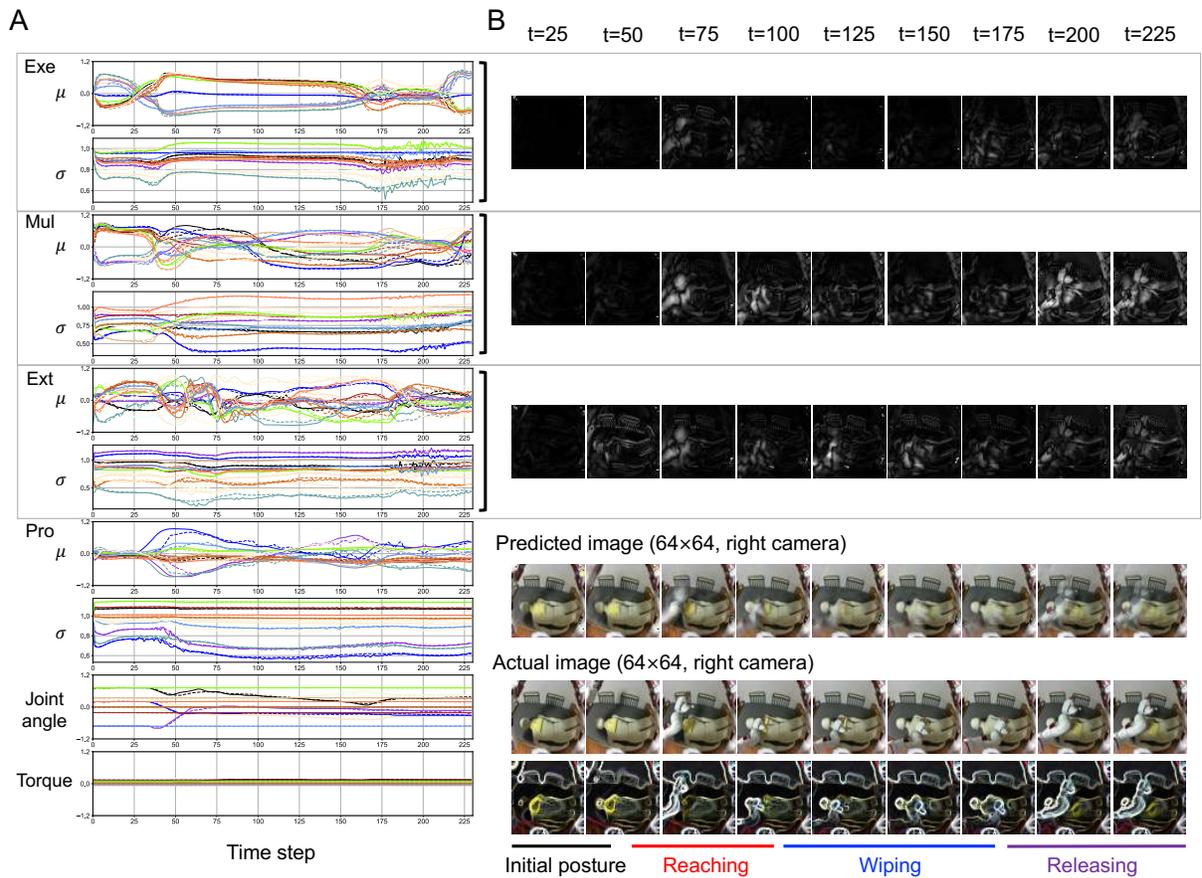

**Fig. 3. Hierarchical probabilistic inference during the wiping task.** (**A**) Dynamics of probabilistic latent states in each module, along with predicted and actual proprioceptive inputs. For the mean $\mu$ and sigma $\sigma$ (standard deviation) of latent states, posteriors and priors are shown as solid and dashed lines, respectively. For joint angles and torques, actual and predicted values are indicated by solid and dashed lines, respectively. Different colors represent distinct latent states, joint angles, or torques. For clarity of presentation, the time series of latent states are plotted for 15 selected states. Exe: Executive module. Mul: Multimodal associative module. Ext: Exteroceptive module. Pro: Proprioceptive module. (**B**) Visual representations in each module, together with predicted and actual visual inputs. For reference, an additional image emphasizing the contours of the actual visual inputs is shown beneath them. Contributions of latent states were analyzed through an ablation study. Brighter regions indicate greater influence of the corresponding module's latent states on generating visual predictions. Results are shown for 64×64 images from the right camera.

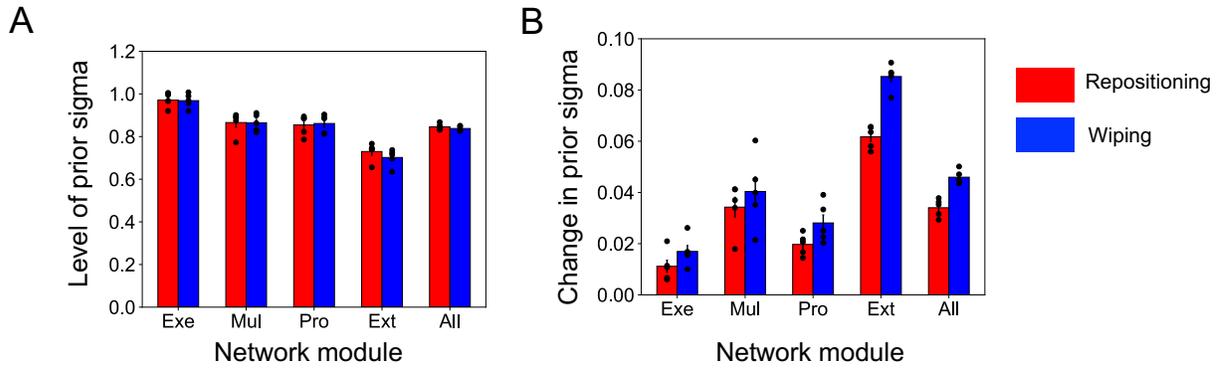

**Fig. 4. Estimated uncertainty and its temporal variability in repositioning and wiping tasks.** (**A**) Mean levels of prior sigma estimated in each module. A paired t-test indicated that the difference in mean prior sigma across all modules between the repositioning and wiping tasks was not significant ($t(4) = 1.67, p = 0.17$). (**B**) Temporal changes in prior sigma, reflecting variability in uncertainty across time. A paired t-test showed that the change in prior sigma averaged across all modules was significantly smaller in the repositioning task than in the wiping task ($t(4) = -8.15, p = 0.0012$). For both (A) and (B), values were first averaged across all time steps and latent states within each trial, and then averaged across 5 trials, 3 test sequences, and 5 independently trained networks for each task. Error bars represent standard errors across the 5 networks. Exe: Executive module. Mul: Multimodal associative module. Ext: Exteroceptive module. Pro: Proprioceptive module. All: All modules. Error bars represent standard errors across the 5 networks.

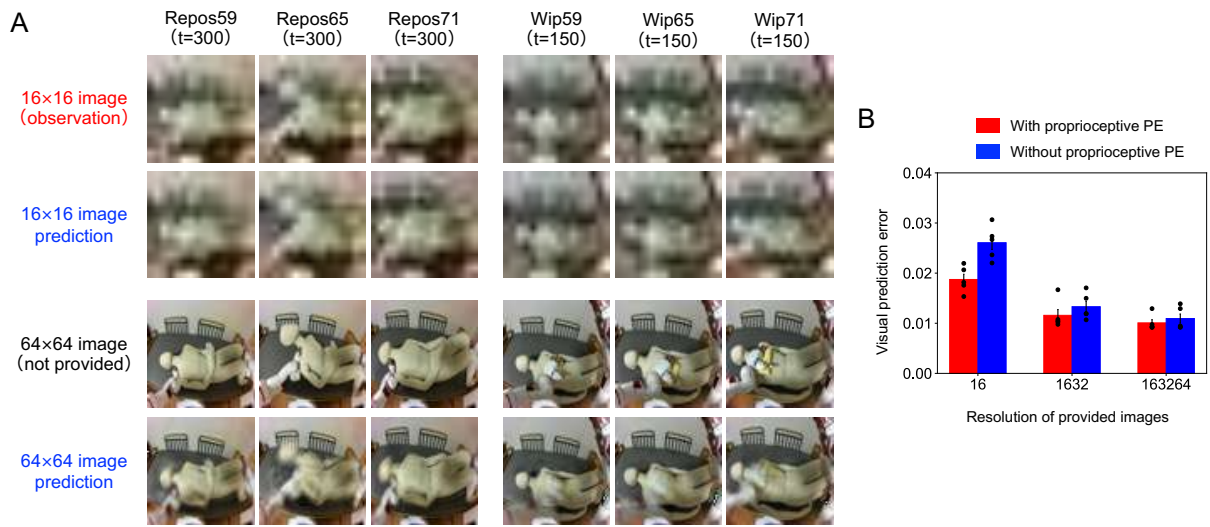

**Fig. 5. Robustness under limited visual inputs.** (**A**) Examples of observed and predicted images at different resolutions. From top to bottom: observed 16×16 image, predicted 16×16 image, non-provided 64×64 image, and predicted 64×64 image. Results are shown for three bed-height conditions (590 mm, 650 mm, and 710 mm) in both tasks. (**B**) Effects of proprioceptive input on visual prediction errors under different image-resolution conditions. Prediction errors were computed across all image resolutions and averaged over 5 trials, 6 test sequences, and 5 independently trained networks, separately for conditions with and without proprioceptive prediction error (PE). Error bars represent standard errors across the 5 networks. A two-way nonparametric repeated-measures ANOVA using the aligned rank transform (ART-ANOVA) revealed significant main effects of image resolution ($F(2,20) = 105.57, p < 0.001$) and proprioceptive condition (with vs. without proprioceptive PE) ($F(1,20) = 46.09, p < 0.001$), as well as a significant interaction between them ($F(2,20) = 13.57, p < 0.001$). Post hoc pairwise comparisons with Holm correction confirmed that prediction errors were significantly lower in the presence of proprioceptive input than in its absence ($p < 0.001$). A significant simple effect of proprioceptive condition was also observed when only 16×16 images were provided ($p = 0.034$).

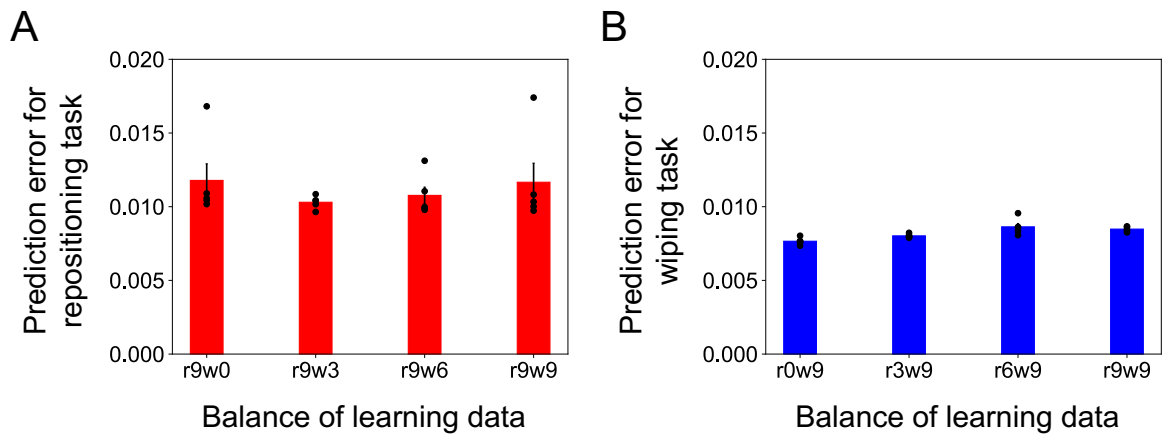

**Fig. 6. Interference between tasks in learning.** (**A**) Prediction errors for the repositioning task under different data-balance conditions. The number of learning sequences for repositioning was fixed at 9, while the number of wiping sequences was varied. A one-way aligned rank transform ANOVA (ART-ANOVA) reported no significant differences in prediction error among data conditions ($F(3,16) = 0.62, p = 0.61$). (**B**) Prediction errors for the wiping task under different data-balance conditions. The number of learning sequences for wiping was fixed at 9, while the number of repositioning sequences was varied. A one-way ART-ANOVA indicated a significant effect of data condition on prediction error ($F(3,16) = 19.16, p < 0.001$). Post hoc multiple comparisons using the Holm correction showed that the prediction error for the wiping task was significantly lower when the number of repositioning sequences was 0 than when it was 6 ($p < 0.001$) or 9 ($p < 0.001$). In addition, prediction error was significantly lower when the number of repositioning sequences was 3 than when it was 6 ($p = 0.0033$) or 9 ($p = 0.0033$). For both (A) and (B), values are averaged across 5 trials, 3 test sequences, and 5 independently trained networks for each learning condition. Error bars indicate standard errors across the 5 networks.